# Understanding Sampling Style Adversarial Search Methods

**Raghuram Ramanujan** and **Ashish Sabharwal** and **Bart Selman**
Department of Computer Science
Cornell University, Ithaca NY 14853-7501, U.S.A.
{raghu,sabhar,selman}@cs.cornell.edu

## Abstract

UCT has recently emerged as an exciting new adversarial reasoning technique based on cleverly balancing exploration and exploitation in a Monte-Carlo sampling setting. It has been particularly successful in the game of Go but the reasons for its success are not well understood and attempts to replicate its success in other domains such as Chess have failed. We provide an in-depth analysis of the potential of UCT in domain-independent settings, in cases where heuristic values are available, and the effect of enhancing random playouts to more informed playouts between two weak minimax players. To provide further insights, we develop synthetic game tree instances and discuss interesting properties of UCT, both empirically and analytically.

## 1 INTRODUCTION

The recent introduction of the Upper Confidence bounds applied to Trees (UCT) method for adversarial game playing significantly improved the standard of computer Go programs (Gelly and Silver, 2007, 2008). In fact, it now appears that we may reach human-level performance in Go within the next decade, which is substantially sooner than anyone had predicted just a few years ago. The current developments are especially surprising given that the traditional minimax game tree search, which has yielded world-class play in Chess and many other games, does not scale to the domain of Go. Two issues hamper the application of minimax search to Go: a very high branching factor and the lack of a high-quality board evaluation function. A good board evaluation function is key in game tree search when one cannot reach terminal states in the game tree. UCT provides an effective way to address both these issues.

The UCT algorithm (Kocsis and Szepesvári, 2006) is derived from a highly effective approach to solving the multi-armed bandit problem called UCB1 (Auer et al., 2002). The UCT sampling strategy strikes a provably optimal balance between exploration of new game states and exploitation, where lines of play that appear promising are repeatedly searched to deeper levels. This novel approach means that UCT can reach regions of the search space that are much deeper than the conventional iterative deepening minimax search, which has been the "gold standard" for Chess and other games. When UCT encounters a non-terminal leaf node, a random (or weakly informed) playout is typically used to provide some indication of the value of the state. As the designers of UCT for Go have observed, it is somewhat counter-intuitive that there is any useful information to be gained from having two weak players play out the game to completion from some intermediate state. After all, any real game between competent players will follow a very different overall trajectory than one between weak players.

In Go, these properties of UCT have been very useful and clearly alleviate some of the difficulties of doing a standard minimax search: the more focused search can go much deeper than any kind of iterative deepening minimax search given the high branching factor of Go, and the playouts provide useful board evaluation information, given that a good general board evaluation function for Go is not known.

The success of UCT in Go raises the natural question of whether UCT is also effective in other adversarial reasoning domains. We address this question by studying UCT in the context of Chess as well as synthetic instances designed to highlight the key aspects of UCT. We chose Chess as one of our evaluation domains mainly because standard minimax search works so well for it. We can therefore study the behavior of UCT and its two key new search concepts in detail by comparing its performance with traditional minimax results as the "gold standard". As we will see,

UCT *per se* is not competitive in Chess. However, there are promising aspects of UCT that may be used to complement more traditional search. We will also identify what causes difficulties for UCT in Chess style domains. Our results are applicable to any adversarial reasoning domain that has the characteristics we identify.

After discussing the basics of UCT in Section 2, we will empirically show in Section 3 that in domain-independent settings, UCT can easily outperform minimax search with a comparable amount of computational power. We will then describe in Section 4 how the performance of UCT can be significantly boosted when heuristic information is available and is used in place of random playouts. The way we use heuristics is much more direct than the "bootstrapping" approach often used to initialize UCT leaf values. However, even with a high-quality heuristic, UCT does not perform well on Chess compared to a shallow minimax search. This suggests that the different success rates of UCT on Chess and Go is perhaps explained not so much by the lack of good heuristics as by the intricate properties of the two underlying search spaces.

We will then return to playouts and demonstrate that playouts between slightly more informed players than random players can lead to discovering information that is available only to a much deeper minimax search. Not being able to discover such information can lead to UCT falling into what we call *soft traps*; we will show that soft traps are pervasive even in grandmaster games of Chess and will provide a concrete example. Finally, we will turn our attention to the case of *synthetic instances* designed to provide insights into the behavior of UCT in practice, complementing known theoretical results about bandit-based sampling and UCT that provide worst case exponential time convergence guarantees in the limit. We focus, in contrast to existing analysis (e.g., Auer et al., 2002; Gelly and Silver, 2007; Coquelin and Munos, 2007), on simple cases such as binary trees with implanted winning strategies of low complexity (i.e., few critical moves) where UCT does work in practice, and provide a methodology to analyze the behavior of UCT on such trees.

This part of the paper highlights and formalizes several subtle aspects of UCT. For example, we show, both empirically and analytically, that the time to convergence scales exponentially with the depth of the critical choice points in a winning strategy. In fact, we provide an expression capturing the fact that the runtime of UCT can be decomposed *additively* into the time it spends identifying certain "active" nodes at the first critical level, then the time it needs to explore the sub-trees from these active nodes to identify active nodes at the next critical level, and so on. We study the "averaging" backup strategy employed by UCT and show how it can make recovering from early poor choices very tricky and expensive. This suggests that other backup strategies may work better, but designing one needs further study. We also allude to differences between single agent search as in UCB1 (Auer et al., 2002), which has been the motivation for the multi-agent UCT algorithm, and multi-agent scenarios. For example, while single agent sampling based search can easily break ties between several good moves and "freeze" to one such good move, in two-player minimax settings, the opponent constantly keeps switching in the hope of finding a better move, thus preventing the search from "freezing" onto a single principal variation. This results in exponential scaling of UCT in two-player games that would not occur in single-agent search.

## 2 BACKGROUND

Monte Carlo sampling techniques have been successfully applied in the past to produce expert-level play in games of incomplete information such as Bridge (Ginsberg, 1999) and Scrabble (Sheppard, 2002). However, they have seldom outperformed traditional adversarial planning techniques such as the minimax algorithm in deterministic 2-player game settings such as Chess. This changed recently with the emergence of UCT, which was used to produce the first program capable of master level play in 9x9 Go (Gelly and Silver, 2007, 2008), a domain which had thus far proven to be challenging for minimax presumably due to a large branching factor and lack of good heuristics. UCT has also proved promising in new domains such as Kriegspiel that were beyond the scope of any traditional planning techniques (Ciancarini and Favini, 2009), and in general game playing (Finnsson and Björnsson, 2008).

For two-player games, a single iteration of UCT starting at a state $s$ comprises the following steps:

**Selection:** The algorithm selects an action $a$ that maximizes an upper confidence bound on the action value: $\pi(s) = \arg\max_a \left( Q(s,a) + c\sqrt{\frac{\log n(s)}{n(s,a)}} \right)$ where $Q(s,a)$ is the current estimate of the value of taking action $a$ at state $s$, $n(s)$ is the total number of visits to state $s$ over past iterations and $n(s,a)$ is the number of times action $a$ was selected in past visits to $s$. If $n(s,a) = 0$ for any action $a$, it is selected before any other actions are re-sampled. The opposing player symmetrically selects an action that minimizes a lower confidence bound. The constant $c$ determines how the agent trades-off exploiting known good moves and exploring under-sampled ones; in our experiments with Chess, this constant was fixed at 0.4 which produced a good balance between the two strategies.

**Estimation:** The selection operator is repeatedly applied until a previously unvisited state is reached. If this state is non-terminal, a *default policy* is typically used to play out the game from the current position to a terminal state with reward $R$ ($R$ could alternately be a heuristic board evaluation) and the new state is added to the tree. Thus, on each iteration, the size of the tree grows by 1 node. In our experiments, the default policy selects uniformly at random from the available actions (unless noted otherwise).

**Value Backup:** The reward $R$ from the current UCT episode is used to update the values of all state-action pairs on the path from the root to the fringe of the tree by incrementing both $n(s)$ and $n(s,a)$ by 1, and incrementing $Q(s,a)$ by $(R-Q(s,a))/n(s,a)$. This update assigns to each state-action pair the *average* reward accrued from every episode that passed through it.

## 3 DOMAIN-INDEPENDENT SETTINGS

We begin by exploring the extent to which UCT-style search methods can compete with minimax search in a fully domain-independent setting. This situation arises, for instance, in reasoning about quantified Boolean formulas (QBF) where all we have as input is a formula, without any information about the semantics of the variables or the specifics of the problem domain the formula is encapsulating. This also happens in the general game playing setting (Finnsson and Björnsson, 2008).

For our empirical exploration of the behavior of UCT and minimax, we use the setting of Chess but modify minimax to avoid using any Chess-specific heuristic information, pretending that the domain is unknown. Specifically, for $k \geq 1$, MM-$k$-R denotes the minimax player that performs a minimax search of depth $k$, uses $\pm 1$ values at a leaf if it corresponds to a terminal state, and uses the outcome of a single random rollout if the leaf corresponds to a non-terminal state. This produces a player that is aware of winning (losing) positions within its search horizon, but otherwise has the same rollout style information as is available to UCT.

**Experimental Setup.** The results are reported in Table 1, which gives the **success rate** of the column player against the row player. The success rate, throughout this paper, is computed by assigning a score of 0 to each game lost, 1 to each game won, and 0.5 to each game that resulted in a draw. Note that if $m$ games are played between two players, the sum of the success rates of the two players will be precisely $m$. Further, if each of players A and A' wins 3/4 of the non-drawn games against B but A draws fewer games, then the success rate of A will be higher than A' — a desirable property. In this and all experiments, unless otherwise stated, we report the average success rate over a total of 100 games played from the default starting position of Chess, with 50 played as White and 50 as Black. The variation amongst the games is induced by the stochastic nature of (at least one of) the players.

Table 1: UCT and a purely Random player compared against minimax without domain knowledge. Table reports the success rate of the column player against the row (minimax) player.

| Minimax | | UCT | Random |
|---|---|---|---|
| depth | #nodes | | |
| MM-2-R | 1,000 | 74% | 6% |
| MM-4-R | 10,000 | 94% | 0% |
| MM-6-R | 200,000 | 96% | 0% |

The players used for comparison are UCT with random playouts (UCT) and the "random" player that simply selects a legal move uniformly at random. The UCT player is given roughly the same amount of computation power, measured using the number of nodes explored (rather than runtime, in order to discount any implementation differences), as the minimax player it is competing against. We observe that even though MM-$k$-R acts without much information in many situations, it is far from a trivial player as evidenced by its clear success against the random player. Also, searching deeper improves the performance of MM-$k$-R; not only is the success rate of MM-6-R against the random player higher than that of MM-2-R, in a direct playoff (not shown in the table), MM-6-R has a success rate of 66% against MM-2-R. Finally, UCT significantly outperforms MM-$k$-R, demonstrating the potential of UCT in completely knowledge-free settings.

## 4 BOOSTING UCT WITH HEURISTIC INFORMATION

We now consider the setting where we do have prior domain knowledge. We are interested in the extent to which this can be exploited to enhance UCT. Heuristics have already provided promising results for Go. Typically the heuristic value is used to initialize the value of leaf nodes to bias the selection process in the early iterations of the search. Nonetheless, since current heuristics in Go are not very strong, UCT is set up to fairly quickly override the heuristic value with playout values once the node has been visited sufficiently many (typically a few dozen) times. In contrast, for Chess, we have heuristics that are much more pow-erful, and we explore how much they can boost the performance of UCT.

To evaluate this, we consider the player UCT-H that uses the board evaluation heuristic of `gnuchess` at the leaves visited by UCT, rather than the $\{-1, 0, +1\}$ values obtained from random playouts; in other words, we fully replace playouts with heuristic evaluations. This still preserves the convergence properties of UCT, i.e., with sufficiently many iterations, UCT-H will converge to the true minimax value of each node. We carefully rescaled the heuristic value to fall in the range $[-1, +1]$ by resetting the default checkmate valuation of `gnuchess` to the observed maximum heuristic value of a non-terminal node $(6, 500)$. Out of other candidate rescaling schemes including sigmoidal functions, this simple scheme worked the best.

Against a UCT player (with random playouts) that was given $10,000$ iterations for convergence, we found that UCT-H had a success rate of $55.0\%$, $85.5\%$, and $96.5\%$ with only $50$, $100$, and $1,000$ iterations, respectively. Thus, not only is UCT-H significantly faster then UCT per iteration (because it does not do playouts and thus avoids relatively expensive repeated move generation), it needs drastically fewer iterations to be competitive with UCT.

A natural question to ask, then, is how well does UCT-H actually compete as a player against minimax? Unfortunately, for games such as Chess where minimax is *the* successful strategy, even UCT-H doesn't fare too well. We found that even with $50,000$ iterations, UCT-H is only about as powerful as MM-2, a 2-level minimax search with the `gnuchess` heuristic. This suggests that the difference in the performance of UCT in Go vs. Chess is not only due to the quality of the heuristic but perhaps more importantly, due to the different nature of the two underlying search spaces and how "winning" is defined in the two settings. Any successful sampling-based player for Chess must therefore take these aspects into account.

## 5 ENHANCING RANDOM PLAYOUTS

We now focus our attention on one of the two key aspects of UCT, *random playouts*, and ask whether such playouts can provide useful information in domains such as Chess where we already have well-designed state evaluation heuristics. An interesting question in the context of playouts is, *is it at all possible to obtain useful information about a strong player by doing several playouts between two weak players?* We find that random playouts tend not to provide any more information than Chess heuristics themselves, but a slightly more powerful playout—namely a playout between two MM-2 players—*can*, surprisingly, reveal information that is often visible only to a significantly deeper minimax player such as MM-8. We quantify this in terms of a strong correlation between move rankings obtained by the two players.

Such information, visible only to relatively deep and systematic minimax searches, can take the form of *traps* as recently studied by us (Ramanujan et al., 2010), where making the "wrong" move leads to a state from which the opponent has a relatively simple winning strategy; such traps, even at surprisingly shallow depths, were found to be abundant even in grandmaster games of Chess. More generally, we consider here the notion of **soft traps**, where a wrong move takes one to a game state from which the opponent has a guaranteed strategy for gaining significant "advantage" in the game. This advantage could be measured in terms of an evaluation function $h$ for the states. In our analysis of 50 complete grandmaster games, we discovered that $52\%$ of them had at least one occurrence of a soft trap, i.e., a position where an MM-8 and MM-2 search had a significant disagreement over the valuation of the best move. We now make the notion of soft traps precise.

As a generalization of $k$-move winning strategies (Ramanujan et al., 2010), consider a heuristic state evaluation function $h$ and a parameter $\Delta$. Define a $k$**-move** $(h, \Delta)$ **advantage strategy** starting from the current state $s$ as a length-$k$ action sequence that results in a board state $s'$ such that $h(s') \geq h(s) + \Delta$. Note that when $\Delta$ is sufficiently large, this becomes a $k$-move winning strategy.

**Definition 1.** Let $G$ be a 2-player game with a heuristic evaluation function $h$, and $\Delta > 0$ be a constant. The current player $p$ at state $s$ of $G$ is said to be **at risk of falling into a soft trap** if there exists a move $m$ from state $s$ such that after executing $m$, the opponent of $p$ has $k$-move $(h, \Delta)$ advantage strategy. The state of the game after executing $m$ is referred to as a **soft level-$k$ search trap** for $p$.

In Figure 1, we explore how good heuristics and various kinds of playouts are in obtaining information that is visible to a strong player, such as a deep MM-$k$ player (for experimental purposes, we use MM-8 as the gold standard). For this evaluation, we consider boards taken from grandmaster games and compute the *ranking* from best to worst (1 being the best) of the possible moves as given by an MM-8 evaluation of each resulting state. Note that during actual gameplay, only the relative ordering of moves matters; it is for this reason that we choose to study the correlation of the move rankings rather than their raw estimated values. This also helps circumvent the problem of com-paring leaf value estimation methods whose outputs do not map to the same range of values. For each kind of estimation method, we apply smoothing by considering estimates within some $\epsilon$ of each other as ties and assigning them the same rank.

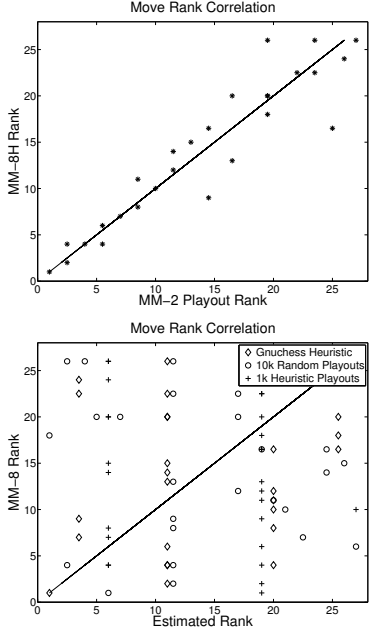

Figure 1: Correlation of move rankings of various players (x-axis) against MM-8 rankings (y-axis). Top: playouts using MM-2. Bottom: `gnuchess` heuristic, random playouts, heuristic playouts.

Figure 1 shows the results for a typical grandmaster board 16 moves (31 plys) deep into the play. In the top panel, we compare for each child, its MM-8 ranking (y-axis) against the ranking obtained based on playouts using two MM-2 players (x-axis). The points being almost on the diagonal shows that the two rankings are very well correlated, especially in the region of most interest—the bottom-left region, representing moves that are considered very good by both players. In contrast, the lower pane of the figure shows that the rankings obtained using the `gnuchess` heuristic, random playouts, or playouts between heuristic players (x-axis) are much more loosely correlated with MM-8 rankings (y-axis). For example, points in the top left corner represent moves that MM-8 thinks are very poor but the other player thinks are quite good—indicative of traps or soft traps missed by the weaker player. Similarly, points in the bottom right corner indicate good moves, as identified by MM-8, that are dismissed as bad moves by the weaker player.

Overall, this demonstrates that playouts between slightly informed players, namely MM-2 players in this case, *can* have a strong correlation with information that is usually visible only to a much stronger player, namely MM-8 in this case. A natural question to ask at this stage, *how does the ranking induced by an MM-2 search itself compare to that induced by a playout between two MM-2 players*? We have discovered that there are in fact situations in which a playout of two MM-2 players uncovers information that an MM-2 search does not. Example 1 describes a concrete occurrence of this phenomenon.

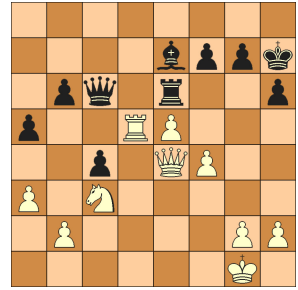

Figure 2: A board where playouts with MM-2 players are able to discover a soft trap visible at depth 9 while complete MM-2 search misses it.

**Example 1.** Consider the Chess board shown in Figure 2. We will follow the standard algebraic chess notation in our discussion, where rows (ranks) are labeled 1-8 and columns (files) are labeled a-h, with a1 being the bottom left corner. In the given state, the Black king is in check with Black on move and an MM-2 search recommends that the king be moved to h8. However, this allows White a devastating countermove: moving its pawn on file f to f5 and thereby trapping Black's rook. Black can stall for two moves by using its bishop to place the White king in check, and subsequently freeing its rook to escape up file e. In this case, White simply moves its own rook to the same rank as the Black rook. This sets up a situation where Black is at minimum forced to trade its queen and rook for the White queen. Sub-optimal sequences of play result in much costlier piece exchanges for Black. The correct move in the original position is for Black to move its pawn on file g to g6, thereby nullifying White's pawn threat—this is the move prescribed by a complete MM-8 search, as well as an MM-2 playout.

## 6 INSIGHTS INTO UCT: SYNTHETIC SEARCH SPACES

While UCT is easy to describe, it has a rich and complex behavior on adversarial search spaces such as those of Chess and Go. In order to better understand its behavior, we consider synthetic adversarial search spaces where we vary, in a controlled manner, key properties that affect the performance of UCT.

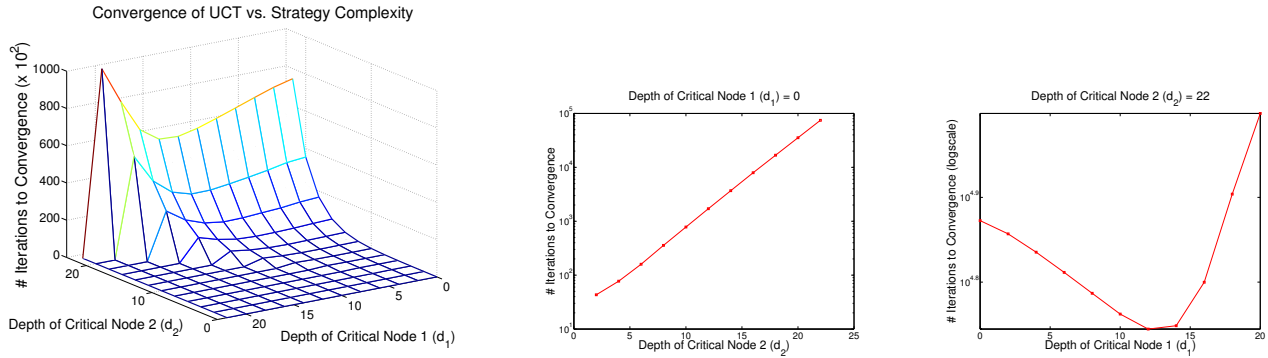

Figure 3: UCT convergence time as a function of the depths of the critical nodes. left: 3-D contour; middle: slice with a fixed depth of critical node 1, in logscale; right: slice with a fixed depth of critical node 2, in logscale.

We study game trees with implanted winning strategies for the max player (denoted Max) who is on move at the root node. The winning strategies are parametrized by the number of *critical decision nodes* and their depths. If Max makes the correct action choice at every critical node, then regardless of the actions chosen by either player at all other nodes, the payoff at the end of the game is $+1$. If Max chooses an incorrect action at any of the critical nodes, then the payoff at the end of the game is drawn uniformly from $\{-1, 0, +1\}$. This simple model captures the notion of winning plans that exist in many tactical games like Chess, where from a given state, a player can force a win by executing a sequence of a few clever moves.

In these experiments, we are interested in the time UCT takes to "discover" the winning strategy for Max, which we define in terms of the utility assigned by UCT to the root node. Once UCT has settled on a winning sequence of moves for Max (i.e., a *principal variation*), it will exploit it on subsequent iterations and this will force the utility of the root node to approach $+1$. A subtle point is that the min player (denoted Min) might keep forcing Max to different principal variations; nonetheless, the paths will be equally good for Max and the value of the root will still approach $+1$.

Formally, let $v(t)$ be the utility assigned to the root node of the search tree after $t$ iterations of UCT. For a single UCT search, we define the $\tau$-**convergence point** $t^*$ as the smallest $t$ such that $v(t) \geq \tau$ for all $t \geq t^*$. We say that UCT has $\tau$-**converged** if the current iteration number is at least $t^*$. Unless otherwise specified, we will simply use the term converged to imply $\tau$-convergence at the root with $\tau = 0.7$.

### 6.1 EMPIRICAL OBSERVATIONS

Figure 3 illustrates how the time UCT takes to converge in the presence of 2-step winning strategies (i.e., strategies with 2 critical nodes) in a 24-level binary tree varies as a function of the depths of the two critical nodes (hereafter referred to as $d_1$ and $d_2$, with $d_2 > d_1$). Note that in the mesh plot, the area of interest lies beyond the $d_1 = d_2$ line, towards the back of the plot. The middle and right-most panels depict slices of this surface obtained by fixing $d_1$ and $d_2$, resp.

As seen in the middle panel, for a fixed $d_1$, the convergence time of UCT is essentially exponential in $d_2$. The dependence of the convergence time on $d_1$ is more intriguing—with a fixed $d_2$, UCT appears to perform best when $d_1$ is slightly more than half of $d_2$. This "dip" in the curve is captured by the following expression for the runtime of UCT, which we explain below:

$$UCT(d_1, d_2) = a \cdot C^{d_1/2} + b \cdot 2^{d_1/2} \cdot C^{(d_2-d_1)/2} \quad (1)$$

where $2 < C < 3$ (empirically 2.37) and $a, b > 0$ are small constants. This expression fits the mesh plot in Figure 3 very closely and highlights a key property of UCT in the presence of multi-step winning strategies: *The runtime of UCT can be decomposed additively into the time spent between consecutive critical levels.* Specifically, UCT first explores roughly $2^{d_1/2}$ "active" nodes at level $d_1$ in time $O(C^{d_1/2})$, then explores each of the roughly $2^{d_1/2}$ subtrees below these active nodes at level $d_1$ in time $O(C^{(d_2-d_1)/2})$ each to identify roughly $2^{(d_2-d_1)/2}$ active nodes at level $d_2$ in each subtree, and so on down to other critical decision levels. The quantity $2^{d_1/2}$ (in general, $2^{(d_i-d_{i-1})/2}$ per subtree) representing "active" nodes is nothing but the minimum number of nodes that Min can continually force Max to explore until Max has figured out a winning sequence from all of these nodes. In general, we can extend this reasoning to $k$ critical decision levels, suggesting that the runtime of UCT is captured by:

$$UCT(d_1, d_2, \ldots, d_k) = \\ a \cdot C^{d_1/2} + b \cdot 2^{d_1/2} \cdot UCT(d_2, \ldots, d_k) \quad (2)$$

Note that Max takes $C^{d_1/2}$ iterations, and not $2^{d_1/2}$, to identify the $2^{d_1/2}$ active nodes at level $d_1$ that Min can force it to. This is because, although Max ideally has the choice to "freeze" to any one of its equally good children, the exploration constant forces Max to explore to some extent the other child as well, especially during the initial few visits to that node. Nevertheless, the overall time is much less than the size of the full search tree till this level, which is $2^{d_1}$ or $4^{d_1/2}$.

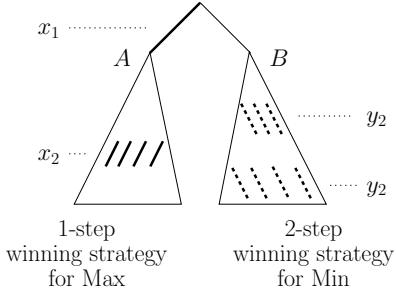

Figure 4: Synthetic binary trees with implanted winning strategies for both Max and Min.

Table 2: Effects of the depths of Max and Min's strategies on UCT's convergence time. 'F' and 'U' denote instances with favorable and unfavorable initial estimates, respectively.

| Max's Strategy | Min's Strategy Depth | | | | | |
| Depth | Shallow | | Mid-level | | Deep | |
| | F | U | F | U | F | U |
| --- | --- | --- | --- | --- | --- | --- |
| Shallow | 36 | 148 | 77 | 610 | 560 | 3800 |
| Mid-level | 950 | 1000 | 1500 | 1900 | 7900 | 13k |
| Deep | 16k | 16k | 17k | 17k | 30k | 33k |

Table 3: Effects of the depths of Max and Min's strategies on the distribution of visits to the right subtree.

| Max's Strategy | Min's Strategy Depth | | | | | |
| Depth | Shallow | | Mid-level | | Deep | |
| | F | U | F | U | F | U |
| --- | --- | --- | --- | --- | --- | --- |
| Shallow | 33% | 21% | 34% | 30% | 40% | 36% |
| Mid-level | 3% | 4% | 14% | 16% | 31% | 31% |
| Deep | 0.2% | 0.3% | 1.8% | 2% | 16% | 16% |

In our second experiment, we study a more complex scenario where both Max and Min have implanted strategies and a few initial samples provide incorrect guidance at the root (see Figure 4). In particular, we study binary trees of depth 20 where Max has critical nodes at depths $(x_1, x_2)$ where $x_1 = 0$, $2 \leq x_2 \leq 18$, and $x_2$ is even, and Min has critical nodes at $(y_1, y_2)$ where $1 \leq y_1, y_2 \leq 19$ and $y_1, y_2$ are odd. In order to win, Max must move left at the root and again at level $x_2$; if Max goes right at the root, then Min can force a win by going left at levels $y_1$ and $y_2$ (i.e., the right child of the root is a trap state for Max). Let $A$ and $B$ be the subtrees rooted at the left and right children of the root node respectively. We bias the values of the leaves that are not on a winning path for either player such that the average of the values of the leaves in $A$ is 0, while the average of the values of the leaves in $B$ is 0.5. Thus, the $B$ subtree, though ultimately a losing proposition for Max (assuming optimal play by Min) will look more promising with limited sampling. We now ask the question, *how do the depths of the strategies for the two players influence UCT's convergence time?*

Table 2 presents our findings based on an average of 100 UCT runs on a fixed tree. On its first few iterations, UCT receives extremely noisy estimates of the utilities of its two children at the root. In the best or "favorable" case, these initial estimates correlate correctly with the true utilities of the children and Max chooses to explore subtree $A$ first. In the unfavorable case, the child rankings are reversed and Max chooses to explore subtree $B$ first. Note that any ties will eventually resolve one way or the other, and at that point, we fall back on one of these two cases.

There are a number of interesting trends in Table 2. First, when estimates are unfavorable at the root, the time to convergence is greater as UCT initially "wastes" time in subtree $B$ until it (at least partially) uncovers Min's winning strategy. Second, this gap in convergence time is most pronounced when either Max has a shallow winning strategy or Min has a deep winning strategy. This too makes sense; in the former case, UCT can uncover Max's strategy very quickly if given the chance, and hence the time "wasted" in subtree $B$ counts relatively much more; in the latter case, UCT simply needs to work harder to uncover Min's winning strategy and switch to subtree $A$.

Finally, we note that increasing the depth of Min's strategy slows down UCT's convergence *even in the favorable instances*. The data in Table 3, which shows the average percentage of time UCT spends in subtree $B$ during the runs presented in Table 2, helps explain this phenomenon. As Min's strategy is implanted deeper down in the tree, UCT spends more time exploring subtree $B$. A by-product of this repeated sampling from $B$ is that the estimated utility of the root node is now heavily biased by the samples drawn from $B$; when UCT eventually switches to subtree $A$ and discovers Max's winning strategy, it needs to work extra hard to overcome this bias and reinforce the true utility of the root (we will formalize this in Section 6.2, equation (3)). This is illustrated by the fact that when both Max and Min have shallow strategies, when UCT converges, the root node of subtree $A$ has a typical utility estimate of 0.720; when both have deep strategies, the root node of subtree $A$ needs to reach a much higher target value of 0.850.

This highlights an important shortcoming of UCT, namely that it can be overly optimistic in its estimates of state utilities, that lead it on wild goose chases. By the time it discovers that an action it has been exploring is sub-optimal, nodes higher up the tree have been reinforced with so many samples that it faces an uphill task in changing these estimates. In the face of computational constraints (for example, in a timed game-playing setting such as Blitz Chess), this is particularly troublesome for it means that UCT could easily have spent its time exploring sub-optimal moves and thus faces a very real risk of falling into a trap state.

## 6.2 ANALYTICAL INSIGHTS

While a few attempts have been made to analyze bandit based sampling methods in general and UCT in particular (e.g., Auer et al., 2002; Gelly and Silver, 2007; Coquelin and Munos, 2007), these analyses are based on the worst case scenario and, in essence, boil down to showing that an exponential (or even super-exponential (Coquelin and Munos, 2007)) number of iterations are necessary and sufficient for UCT to converge to true minimax values. These exponential time convergence results, while intricate and interesting, do not explain the success of UCT in practice in domains such as Go, with a practically limited number of iterations available during game play. In contrast, our goal in this section is to provide a methodology for analyzing some simple scenarios where UCT does work, and obtain insights into its runtime behavior. Specifically, we will consider 2-step winning strategies implanted in binary trees.

We highlight three take-away messages, some of which have previously been observed empirically and are derived here analytically: (a) the averaging backups of UCT can make recovering from poor early choices very costly; (b) UCT in two-player settings scales exponentially with the depth of the critical choice points, whereas in single-player settings, all that matters is the number of critical choice points, not their depth; and (c) the tension between exploration and exploitation as controlled by the exploration constant.

In order to make the analysis easier while still retaining the key aspects of UCT, we work with a modified version of the algorithm in this section. Instead of implementing the UCB1 exploration-exploitation strategy, we will use an $\epsilon$-*greedy* version of the algorithm, where $\epsilon \in [0, 1]$ is a constant determining how often sub-optimal moves are explored. Specifically, when exploring a node for the first few times, UCT simply visits all children once (a "round"), as usual. However, after this round, it selects an optimal branch (breaking ties at random) with probability $1 - \epsilon$ and a sub-optimal branch (breaking ties at random) with probability $\epsilon$. Auer et al. (2002) showed that this simpler variant of UCB1 also has similar good convergence properties (in the limit), as long as $\epsilon$ decreases linearly with the number of times the node is visited.

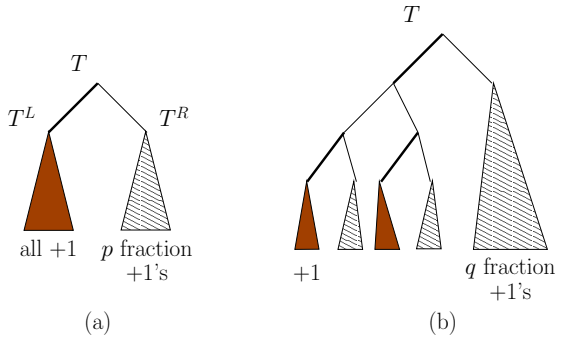

Figure 5: Synthetic binary trees with implanted winning strategies for Max. (a) 1-step winning strategy. (b) 2-step winning strategy.

We make one further modification, where instead of dealing with tie-breaking, we assume that rounds similar to the first round are repeated (i.e., all children explored in each round) until ties are broken. For binary trees, which will be the main focus of this section, this modification does not make a significant difference.

### 6.2.1 Scenario A

For ease of illustration, we start with the simplest case and build upon it. Consider a binary game tree $T$ with Max on play at the top node. Let $T^L$ and $T^R$ denote the left and right subtrees, respectively, of $T$. Suppose that all leaves of $T^L$ are labeled $+1$, i.e., Max has a sure win if he makes the left move. Suppose also that a $p$ fraction of the leaves, where $p \in [0, 1)$, of $T^R$ are labeled $+1$ and the rest are labeled $-1$. This tree is depicted in Figure 5(a), with bold edges corresponding to winning strategy moves. *How long does it take for UCT to identify the left branch as the winning move?*

In a given round at the root node of $T$, a playout from the left child always leads to $+1$ while a playout from the right child leads to $+1$ with probability $p$. Therefore, we have a tie with probability $p$ and it follows that the expected number of rounds needed to break the tie is $1/(1-p)$. Hence, the total number of visits needed to the root node of $T$ in expectation equals $2/(1-p)$ (as there are 2 visits per round) plus the time it takes for UCT to converge at the left child after the tie is broken. Note that the only way for the tie to be broken in this tree is to have all $+1$ playouts on the left and exactly one −1 playout on the right, implying that the left move will necessarily be identified as the optimal move when the tie is broken. (This will not be the case in general, as we discuss later.) From this point on, $T^L$ will be visited a $1 - \epsilon$ fraction of the times the root node of $T$ is visited. Let $C(\tau, value, iter)$ denote the number of visits needed to the winning strategy node (in this case the root node of $T^L$) for UCT to $\tau$-converge at the root node of $T$, where *value* denotes the current value of the node and *iter* denotes the number of visits already made to the node; due to the "averaging" backups of UCT, the current state of the node significantly affects the time to convergence even after a winning strategy has been identified, and we will quantify this shortly. The number of visits needed to the root of $T$ is therefore roughly $2/(1-p) + C(\tau, value, iter)/(1-\epsilon)$.

How do we determine $C(\tau, value, iter)$? In the unlikely case that the current value, *value*, is already at least as good as $\tau$ (i.e., $value \geq \tau$ for Max), this quantity is 0. Otherwise, assuming subsequent visits explore the identified winning strategy, resulting in all +1 playout values, the averaging nature of backups dictates that:

$$\frac{C(\tau, value, iter) \times 1) + (iter \times value)}{C(\tau, value, iter) + iter} = \tau$$

$$\implies C(\tau, value, iter) = iter \times \frac{\tau - value}{1 - \tau} \quad (3)$$

In our case, $iter \approx 1/(1-p)$ and $value = 1 - 2/iter$ as all but the very last round should result in playout values of +1. Plugging these values in, the number of visits to the root node of $T$ till convergence is roughly:

$$\frac{2}{1-p} + \frac{1+\tau-2p}{(1-\epsilon)(1-p)(1-\tau)}$$

**Remark 1.** Equation (3) points out an interesting limitation of UCT that we have already encountered near the end of Section 6.1, namely, that the *averaging backups of UCT can make recovering from poor early choices very expensive*. In particular, if *iter* is high and *value* is too low (for Max), then UCT will take a long time to make up for its mistakes before it reaches $\tau$. This suggests there might be other backup strategies, although finding an effective alternative backup strategy requires further study because natural choices such as simple minimaxing tend to be very brittle.

### 6.2.2 Scenario B

We now explore the "tension" between having a small value for the exploration constant, $\epsilon$, and a large value. This example will also illustrate that the depth of the critical nodes of a winning strategy *exponentially influences* the number of iterations needed for convergence.

This is in stark contrast to $k$-step winning strategies in *single-player settings*, where it is easy to argue that the depth of the critical choice points is immaterial, and all that matters is the number of critical choice points. Intuitively, the difference between the single player and two players settings is that in the former case, since all choices look equally good (or bad) at non-critical points, the player can arbitrarily "freeze" on one of them and keep exploiting it, while in the two player setting, the opponent prevents this freezing by continually forcing the winning player to different areas of the search space in the hope of avoiding defeat. For example, for a depth $d$ winning strategy, the losing player can force the other player to explore precisely $2^{d/2}$ paths.

Suppose that $T$ is modified so that the strategy embodied by $T^L$ in Scenario A is actually hidden deeper and that Max needs to make one good move to get to this strategy. Specifically, we now have a 2-step winning strategy for Max, with critical moves at levels 0 and 2, with the subtrees at level 2 being identical to the ones in Scenario A. Also, let us suppose that the right subtree of the root node has a fraction $q$ of +1 leaves, which will affect tie breaking at the root. This is depicted in Figure 5(b).

Given the expression derived above for Figure 5(a) for the number of times we need to visit each of these subtrees at level 2 in order to identify the winning strategy from there on, how many times do we need to visit the root node of the tree to achieve this? First, consider a node $X$ one level above a winning strategy at level 2. Min is on move at $X$, which means that as soon as Max begins to identify the winning strategy on the left branch of $X$, Min has an incentive to switch to the right branch of $X$ (i.e., what's good for Max is bad for Min). In other words, *Min will keep switching* between the two choices until Max has figured out the winning strategy under both choices of Min. This means that the number of visits to $X$ that we need is *twice* the number of visits to each of $T^L$; in general, when Max's winning strategy is at depth $d$, the number of visits needed will be $2^d$ times the number of visits to any single "winning" subtree at level $d$—hence the exponential scaling with the depth of the winning strategy.

Further, the tie at the root node of $T$ may now be broken in favor of the right child as well, as there are leaves labeled −1 on both sides. If the tie breaks in favor of the left child (the "favorable" case), then the number of iterations needed after breaking the tie is:

$$D(favorable) \approx \frac{2}{1-\epsilon} \times \left( \frac{2}{1-p} + \frac{C(\tau, value, iter)}{1-\epsilon} \right)$$

the latter part of which is similar to Scenario A, mul-tiplied by 2 for twice the work that needs to be done due to Min's choice at level 2, and divided by $(1-\epsilon)$ since in the favorable case we will visit the left subtree of $T$ this fraction of the times we visit the root node of $T$.

More interestingly, when the tie at the root is incorrectly broken in favor of the right hand side child at the root (the "unfavorable" case), the left subtree is visited only an $\epsilon$ fraction of the time, implying that many more visits to the root node are needed in order to achieve the same number of visits as before to the strategy nodes at level 2. Specifically, the number of iterations needed after breaking the tie is:

$$D(unfavorable) = \frac{2}{\epsilon} \times \left( \frac{2}{1-p} + \frac{C(\tau, value, iter)}{1-\epsilon} \right)$$

This illustrates, in a concrete fashion, the tension between small and large values of $\epsilon$, when the goal is to minimize the number of visits to the root node to achieve convergence; see Figure 6 for an illustration where $p = 0.5$ and the $C$ value is taken to be 10.

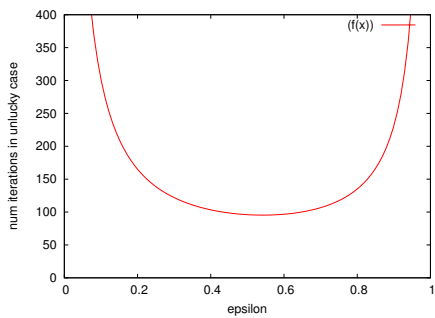

Figure 6: The effect of varying $\epsilon$ on convergence time.

Additionally, we must consider the time to break the tie at the root node, which is slightly more complex than in Scenario A. The fraction of +1 labeled leaves on the right is $q$ and on the left is $p' = 3p/4$. Therefore, the probability of a tie is $p'q$ (when both playouts yield +1) plus $(1-p')(1-q)$ (when both playouts yield −1), giving $p'+q-2p'q$. Thus, the expected number of visits before the tie is broken is $2/(p' + q - 2p'q)$. Further, when this happens, the tie is broken in favor of the left subtree with probability $p'(1 - q)$ and in favor of the right subtree with probability $(1 - p')q$. Putting all this together, we have the following expression for the rough number of visits needed to the root node:

$$\frac{2}{p' + q - 2p'q} + \frac{p'(1-q) \times D(favorable)}{p' + q - 2p'q}$$
$$+ \frac{(1-p')q \times D(unfavorable)}{p' + q - 2p'q}$$

## 7 CONCLUSION

This work provides insights into the behavior of UCT and extends its analysis to complement known worst case (super-)exponential convergence results. We studied UCT in domains such as Chess where traditional minimax search is very effective. Our results demonstrate that UCT consistently beats minimax in domain-independent settings, that it can be significantly boosted by incorporating a state evaluation function, and that more informed playouts can enhance performance. Finally, our results on synthetic instances with implanted strategies revealed an interesting pattern in the convergence behavior of UCT.

### Acknowledgments

Supported by NSF (Expeditions in Computing award for Computational Sustainability, 0832782; IIS grant 0514429) and IISI, Cornell Univ. (AFOSR grant FA9550-04-1-0151).